\documentclass[10pt,journal,compsoc]{IEEEtran}
\ifCLASSOPTIONcompsoc
  \usepackage[nocompress]{cite}
\else
  \usepackage{cite}
\fi
\usepackage[justification=centering]{caption}
\ifCLASSINFOpdf
  \usepackage[pdftex]{graphicx}
  \graphicspath{{Figures/}}
\else
\fi
\begin{document}
%
\title{Artificial Intelligence and its Role in Near Future}
%
%
%
%


\author
{Jahanzaib~Shabbir, and
 Tarique~Anwer}

\markboth{Journal of \LaTeX\ Class Files,~Vol.~14, No.~8, August~2015}%
{Shell \MakeLowercase{\textit{et al.}}: Bare Demo of IEEEtran.cls for Computer Society Journals}
%



\IEEEtitleabstractindextext{%
\begin{abstract}
AI technology has long history which is actively and constantly changing and growing. It focuses on intelligent agents, which contains devices that perceives environment and based on which takes actions in order to maximize goal success chances. In this paper, we will explain the modern AI basics and various representative applications of AI.  In context of modern digitalized world, Artificial Intelligence (AI) is the property of machines, computer programs and systems to perform the intellectual and creative functions of a person, independently find ways to solve problems, be able to draw conclusions and make decisions. Most artificial intelligence systems have the ability to learn, which allows people to improve their performance over time. The recent research on AI tools, including machine learning, deep learning and predictive analysis intended toward increasing the planning, learning, reasoning, thinking and action taking ability \cite{2018arXiv180307608S}. Based on which, the proposed research intended towards exploring on how the human intelligence differs from the artificial intelligence \cite{neisser1996intelligence}.  In addition, on how and in what way, the current artificial intelligence is clever than the human beings. Moreover, we critically analyze what the state-of-the art AI of today is capable of doing, why it still cannot reach human level intelligence and what are the open challenges existing in front of AI to reach and outperform human level of intelligence. Furthermore, it will explore the future predictions for artificial intelligence and based on which potential solution will be recommended to solve it within next decades \cite{neisser1996intelligence}.
\end{abstract}

}

\maketitle

\IEEEdisplaynontitleabstractindextext

%
\IEEEpeerreviewmaketitle

\IEEEraisesectionheading{\section{Introduction}\label{sec:introduction}}

%
%
%
%

 

The term intelligence refers to the ability to acquire and apply different skills and knowledge to solve a given problem. In addition, intelligence is also concerned with the use of general mental capability to solve, reason, and learning various situations \cite{feuerstein2002dynamic,milford2015sequence,fragkiadaki2015recurrent,niekum2015learning,devin2017learning,finn2016deep,rusu2016sim,mohamed2015variational,zhu2017target,cruz2015interactive,vinciarelli2015open,doshi2015deep,wang2015designing,cuayahuitl2015strategic,lake2017building,ohn2016looking,wei2017robotic,mathieu2015deep,chen2015combining,wulff2015efficient,ruiz2015scene}. Intelligence is integrated with various cognitive functions such as; language, attention, planning, memory, perception. The evolution of intelligence can basically is studied about in the last ten years. Intelligence involves both Human and Artificial Intelligence . In this case, critical human intelligence is concerned with solving problems, reasoning and learning. Furthermore, humans have simple complex behaviors which they can easily learn in their entire life \cite{gottfredson1998general}.

\section{Which of these and in what level can today’s artificial intelligence do?}

Today’s Artificial Intelligence (robotics) has the capabilities to imitate human intelligence, performing various tasks that require thinking and learning, solve problems and make various decisions. Artificial Intelligence software or programs that are inserted into robots, computers, or other related systems which them necessary thinking ability \cite{zhang2016technology}. However, much of the current Artificial Intelligence systems (robotics) are still under debate as they still need more research on their way of solving tasks. Therefore Artificial Intelligence machines or systems should be in position to perform the required tasks by without exercising errors. In addition, Robotics should be in position to perform various tasks without any human control or assistance \cite{gottfredson1998general}.
Today’s artificial intelligence such as robotic cars are highly progressing with high performance capabilities such as controlling traffic, minimizing their speed, making  from self-driving cars to the SIRI, the artificial intelligence is rapidly progressing \cite{turan2017non1}. The current attention towards portraying the artificial intelligence in robots for developing the human-like characteristics considerably increases the human dependence towards the technology.  In addition, the artificial intelligence (AI) ability towards effectively performing every narrower and cognitive task considerably increases the people’s dependence towards the technology \cite{parkes2015economic}. Artificial intelligence (AI) tools having the ability to process huge amounts of data by computers can give those who control them and analyze all the information. Today, this considerably increases the threat which makes someone's ability to extract and analyze data in a massive way \cite{gottfredson1998general}.
Recently, Artificial intelligence is reflected as the artificial representation of human brain which tries to simulate their learning process with the aim of mimicking the human brain power. It is necessary to reassure everyone that artificial intelligence equal to that of human brain which is unable to be created \cite{zhang2016technology}. Till now, we operate only part of our capabilities. As currently, the level of knowledge is rapidly developing, it takes only a part of the human brain. As the potential of human brain is incommensurably higher than we can now imagine and prove. Within human brain, there are approximately 100 trillion electrically conducting cells or neurons, which provide an incredible computing power to perform the tasks rapidly and efficiently. It is analyzed from the research that till now computer has the ability to perform the tasks of multiplication of 134,341 by 989,999 in an efficient manner but still unable to perform the things like the learning and changing the understanding of world and recognition of human faces \cite{wong2016artificial}.

\section{Where does the human intelligence differ from AI?}

Artificial intelligence refers to the potential of computer controlled machines/robots towards performing tasks that that almost or similar to human beings. In this case, Artificial intelligence is used to develop various robots that have human intellectual characteristics, behaviors, learning from past experience, have abilities to sense, and abilities to making predications and determine meaning of certain situation \cite{turan2017deep}. Robotic technology is largely trending in the current life which has gained popularity in various sectors such as industries, hospitals, schools, military, music, gaming, quantum science and many others \cite{wong2016artificial}.
Artificial Intelligence is an efficient means that make computers and software control robotic thinking with expert systems that significantly illustrate the intelligent behavior, learning and effectively advice users. In general, AI is basically known as the ability or potential of robotics to decide, solve problems and reason \cite{turan2017endo}. There are various innovations of Artificial Intelligence, for example robotic cars which don’t require a driver to control or supervise them. In addition, artificially intelligent technology (robots) involves smart machines that process a large amount of data that a human being can’t be in position to perform. By so robotics are assuming repetitive duties that require creativity and knowledge base. Furthermore, Artificial Intelligence (AI) is the combination of various technologies that give chance to robotics to understand, learn, perceive or complete human activities on their own \cite{neisser1996intelligence}. 
In this case, Artificial Intelligence programs (robots) are built for a specific purpose such as learning, acting and understating whereas humans intelligence is basically concerned with various abilities of multitasking. In general, an Artificial Intelligence tool is majorly concerned with emphasizing robotics which portrays human behaviors. But however, Artificial Intelligence may fail out at some points due to differences in human brain and computers. In brief, Artificial Intelligence has the potential to mimic human character or behaviors \cite{turan2017deep1}. Furthermore, Artificial intelligence is currently partially developed without advanced abilities to learn on their own but instead given commands to act on.  This will be the ultimate future of artificial intelligence, where the AI machines will be recognized the human behavior and emotions and will train their kernel as per it \cite{martinez2015artificial}.

\section{Why can’t we tell that today’s AI is as clever as human beings? }

Generally, there are various paths towards building the intelligent machines that enables the humans to build the super-intelligent machines and provide ability to machines towards redesigning their own programming in order to increase their intelligence level, which is usually considered as the intelligence explosion. In contrast, the shielded human hunt is basically the emotion. The breakthrough of AI technology can frighten the humanity in a way that machine are unable to effectively transmit the emotions. So, there may be possibility that AI can support us with the tasks and functions which usually not involves the feelings and emotion. Till now, AI machines are not able to control their process, for which they need the intelligence and mind of human beings \cite{martinez2015artificial}. But AI development with same pace may cause threat to the humanity, because the self-learning ability may cause the AI machines to learn destructive things, which may cause killing of humanity in a drastic way. In general, there exist various characteristics which distinguish human level intelligence with Artificial intelligence and they include the following;

Thinking ability; it can be both positive as well as negative because of having emotions, which are not with AI machines. The lack of machine’s emotion may lead to destructive in a situation where emotions are required. Russel Stuart believes that machines would be able to think in a weak manner. In general, there are things that computers cannot do, regardless of how they are programmed and certain ways of designing intelligent programs are doomed to failure sooner or later \cite{weston2015towards}. Therefore, most accurate idea would be to think that it is never going to make the machines have a thought at least similar to human The lack of thinking ability of machines may cause lack in passing the behavioral test. What was later called the Turing Test, proposed that a machine be able to converse before an interrogation for five minutes for the year 2000 and in fact, it was partly achieved. It is concluded then, that the machines can actually think, although they can never have a sense of humor, fall in love, learn from experience, know how to distinguish the good from the bad and other attitudes of the human. Artificial Intelligence: A Modern Approach dedicates its last chapter to wonder what would happen if machines capable of thinking were conceived \cite{turan2018magnetic}.
Then it is when we ask ourselves if it is convenient to continue with this project, take risks and follow an unknown path and believe that what may happen will not be negative. Russel and Norvig believe that AI machines role are to be more optimistic. They believe that intelligent machines are capable of "improving the material circumstances in which human life unfolds" and that in no way can they affect our quality of life negatively.
Reasoning; algorithms mimic the phased reasoning that people use to solve puzzles and guide logical conclusions. For complex tasks, algorithms may require huge computational resources; most of them experience 'combinatorial explosions'. Memory or required computer time will be astronomical with tasks of a certain size. Finding a more effective algorithm to solve the problem is a top priority. Humans usually use quick and intuitive judgment rather than gradual deductions modeled by earlier AI studies. I made progress using the "sub-symbol" solution to the problem. The materialized approach of the agent highlights the importance of sensory - motor skills for higher reasoning. The search of the neural network attempts to imitate the structure in the brain that generates this skill. The statistical approach to AI mimics human guessing abilities.
Planning ability; the objection of those who defend human intelligence against that of machines is based on the fact that they do not possess creativity or consciousness. Planning and creativity could be defined as the ability to combine the elements at our disposal to give an efficient, or beautiful, or shrewd solution to a problem we are facing. That is, we call creativity what we have not yet been able to explain and reproduce mechanically in our behavior. However, the functioning of artificial neural networks can also be considered creative but little predictable and plan able \cite{turan2017fully}.
Action taking ability; the action taking ability of humans are based on emotions, deep thinking and its comparison with how much it is beneficial for the human beings while AI only takes actions either based on their coding, therefore, the lack of emotions and comparison about good and bad increases threats. AI machines are actually very intellectually limited, although they may become brilliant in specialization. A child in his earliest childhood is able to learn to put the triangle in the hollow of the triangle of the toy, can recognize the sounds of animals and begin to apply what he learned on a stage in different ones. Machines are not able to do this at the same time without being previously trained for each task. Generally, there is no neural network in the world capable of identifying objects, images, sounds and playing video games at the same time as people \cite{turan2018deep}. Its limitations are obvious even within the same field of action: when Google's Deep Mind created a system to pass Atari video games, relatively simple for a computing entity due to its two-dimensional progress based on reflexes and trial and error, its neural networks they had to be trained every time a game was completed. Machines are not capable of transferring what they learn to other scenarios that catch them off guard as well as humans can adapt, make use of logic, creativity, ingenuity and reason in any situation, however strange and new they may be \cite{turan2018unsupervised}.
%
%
\begin{figure*}
	\centering
	\includegraphics[width=5in,height=4in]{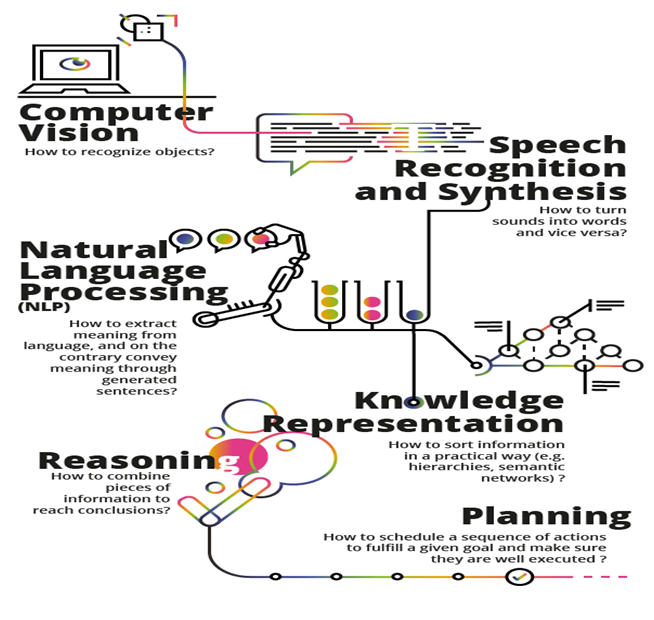}
	\caption{Abilities of Artificial Intelligence}
\end{figure*}

Knowledge representation; it includes the problems machines which expresses the relationships between objects, properties, categories, objects, situations, events, states, and times. The potential cause and effect of knowledge representation is based on what we know about what others know of many other well-studied domains. The concept of "what is present" is an ontology that officially describes a set of objects, relationships, concepts, and properties so that the software’s agent can significantly interpret it. The semantics of this data are recognized as logic to describe roles and descriptions which usually realized as ontology web language classes, properties, and individuals. The most common ontology is called a top ontology that provides the basis of all other knowledge and acts as an intermediary between domain ontology covering specific knowledge about a specific knowledge field. Such formal knowledge representation is based on knowledge using content-based indexing and searching, interpretation of scenarios, support for clinical decisions, automatic reasoning \cite{penfield2015mystery}.
Perception; the machine perception is the ability to derive aspects of the world using input signals from cameras, microphones, sensors, sonar etc., while the computer vision is the ability to analyze visual input.
Learning ability; the development of new algorithm considerably helps the AI machines to learn to write like humans and is able to recognize and draw very simple visual concepts. Generally, the main virtue of human beings is their speed and diversity, when it comes to learning new concepts and applying them in new situations. Computers usually have hard time generalizing from particular samples. In addition, to assessing the ability of the program to learn concepts, they asked people to reproduce a series of characters that had also been plotted by the machine \cite{penfield2015mystery}. Then, they compared them and asked different people what they thought had been done by humans and what by a program, in an adaptation of the classic Turing test that the researchers call visual Turing test. Basically the Turing test is that someone in a room is asking questions to determine if the answers and interactions they receive come from a person or a machine. Therefore, machine learning ability machines significantly makes it much better than that of human being  \cite{penfield2015mystery}.
The natural language process system; they give machines the ability to read and understand human language. A sufficiently powerful natural language processing system enables the use of user interfaces in natural language and gains direct knowledge from written sources such as news texts. Some simple natural language processing applications include information retrieval, text mining, machine translation etc \cite{swain2014approaching}.
The general way to extract and process the natural language, it is essential to use semantic indexing. These indices are expected to increase efficiency as the user speeds up the processor and lower the cost of storing the data despite the user's input being large \cite{parkes2015economic}.

\section{Give a comprehensive survey of what can today’s AI subfilelds do?}

Today’s artificial intelligence is creeping into our daily lives by using the GPS navigation and check-scanning machines. The use of artificial intelligence (AI) in business contributes to the potentialization of various areas of daily life such as customer service, finance, sales and marketing, administration and technical processes in various sectors. Undoubtedly, over the next few years, digital efforts will no longer be isolated projects or initiatives in companies, but the adoption of technologies such as AI at all levels and processes of companies will be a reality to increase their competitiveness. AI begins to integrate into the activities of business. It is important to consider that it has not arrived to replace human tasks, but to complement them and allow people to develop their potential and creativity to the maximum. Introduction of new technologies is a tool for the prevention and fight against corruption, the traceability of electronic actions, and the security that surrounds their management favors confidence in management. This essay will prove how artificial intelligence can improve efficiency of people, help create jobs, and begin to make our society safer for our children. 
Currently, massive research on artificial intelligence significantly improving the world, in which majority of the tasks was performed by the machines, while the role of humans will be to control them. This leads towards the question that is artificial intelligence exceeds the performance level of humans and provide human task in an efficient, quicker and economical level? The paper is about the present and future of artificial intelligence technology and compares it with human intelligence \cite{swain2014approaching}.
The role of this paper is to explore where we are today and what will be the ultimate future of AI technology, if we continuously applied in every field. Further, we will analyses the potential AI techniques and their potential benefits for improving the system.

\section{Briefly explain technical background as well?}

Artificial Intelligence has facilitated us in almost every field of life and has immense scope in future for more productivity and betterment. The origin of artificial intelligence goes back to the advances made by Alan Turing during World War II in the decoding of messages.  The term as such was first used in 1950, but it was only in the 1980s when research began to grow with the resolution of algebra equations and analysis of texts in different languages. The definitive takeoff of Artificial intelligence has come in the last decade with the growth of the internet and the power of microprocessors \cite{renzi2014review}. "Artificial intelligence may be the most disturbing technology the world has ever seen since the industrial revolution" Paul Daugherty, Accenture's chief technology officer, recently wrote in an article published by the World Economic Forum. “This field is now booming due to the increase in ubiquitous computing, low-cost cloud services, new algorithms and other innovations,” adds Daugherty. Developments in Artificial Intelligence go hand in hand with the development of processors that over time have made them start to see these technologies as intellectual, even changing our idea of intellect and forthcoming the perceptions of 'machine', traditionally unintelligent capacity previously assigned exclusively to man.
The AI was introduced to the scientific community in 1950 by the English Alan Turing in his article "Computational Machinery and Intelligence." Although research on the design and capabilities of computers began some time ago, it was not until Turing's article appeared that the idea of an intelligent machine captured the attention of scientists. The work of Turing, who died prematurely, was continued in the United States by John Von Neumann during the 1950s \cite{nilsson2014principles}. His central contribution was the idea that computers should be designed using the human brain as a model. Von Neumann was the first to anthropomorphize the language and conception of computing when speaking of memory, sensors etc. of computers. He built a series of machines using what in the early fifties was known about the human brain, and designed the first programs stored in the memory of a computer \cite{parkes2015economic}.
%
%


\begin{figure*}
	\centering
	\includegraphics[width=6in,height=3.5in]{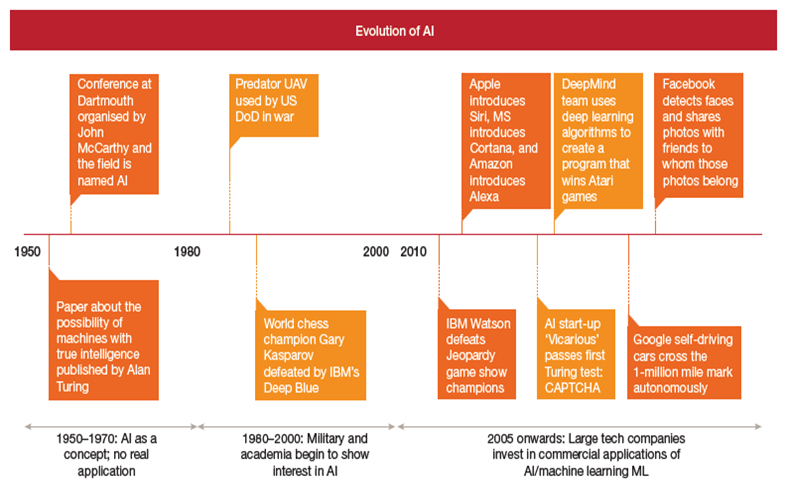}
	\caption{Timeline evolution of Artificial Intelligence}
\end{figure*}

McCulloch (1950) formulate a radically different position by arguing that the laws governing thought must be sought between the rules that govern information and not between those that govern matter. This idea opened great possibilities for AI. In addition, Minsky (1959) modified his position and argued that imitation of the brain at the cellular level should be abandoned. The basic presuppositions of the theoretical core of the AI were emphasis on recognition of thought that can occur outside the brain \cite{gupta2016review,lecun2015deep,papernot2016limitation,levine2016end}. On 1958, Shaw and Simon design the first intelligent program based on their information processing model. This model of Newell, Shaw and Simon was soon to become the dominant theory in cognitive psychology. At the end of the 19th century, sufficiently powerful formal logics were obtained and by the middle of the 20th century, machines capable of make use of such logics and solution algorithms.

\section{What is missing to today’s AI still to be called human level intelligence?}

Humans are different from Artificial Intelligence machine physically in a sense that human race usually experiences the same physical features while the machines takes several forms and shapes. The trans-humanist vision analysis exhibits us to believe that brains are principally the computers. AI reports are the silicon based machines, which was controlled by using the algorithm that reinforces entire internet business. AI believes that once the computers have adequate advanced algorithms, then they will be capable to replicate and enhance the human mind \cite{varela2017embodied}. The tests which exhibits how much AI is distinct from the human intellectual are as follows
Turing test; in order to evaluate on what intelligence means and on how the machine intelligence is different than the human intelligence, the Turing test strongly provide the essential insights to the AI field, which emphasis on how the machine simulates the human thinking. The algorithmic aspects of AI tools should pass the Turning Test \cite{russell2016artificial}. This algorithm will not essentially result in the AGI but may also lean towards applied artificial intelligence. The algorithm tuned through Turing process can also significantly define and passed it.

Eugene Goostman Test; Goostman tests the Turing test and concluded that it has 33\% fooling, which intended him to propose test that should rely on AI in order to efficiently solve the particular tasks that was quite near to AGI. He also concluded that it is not possible for AI machines to be much efficient than that of human because their always acts human actors to work the AI machine. 

\begin{figure}[!t]
	\centering
	\includegraphics[width=2.5in]{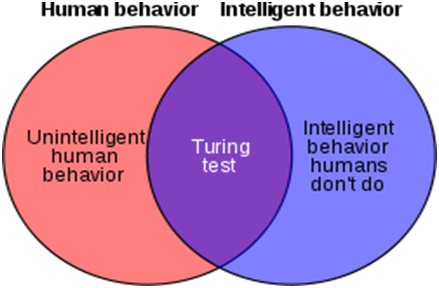}
	\caption{Turing test}
	\label{fig3}
\end{figure}

\section{Importance of Artificial Intelligence}

Artificial Intelligence will revolutionize the way in which different companies across compete and grow across the world by representing a new production factor that can drive business profitability \cite{kang2016dynamic,campos2015diving, russell2015research, milford2015sequence,fragkiadaki2015recurrent,niekum2015learning}. In order to realize the opportunity of AI, most the companies in the world are already developing actively in various Artificial Intelligence strategies \cite{kang2016dynamic}.In addition, they should focus on developing responsible AI systems aligned with ethical and moral values that lead to positive feedback and empower people to do what they know best such as innovation \cite{turan2018sparse}.
With the introduction of successfully implemented Artificial Intelligence (AI) solutions, many industries across the earth can benefit from increased profitability and still count on economic growth. To capitalize on this opportunity, the study identifies eight strategies for the successful implementation of AI that focuses on adopting a human-centric approach and taking innovative and responsible measures for the application of technology to companies and organizations in the world \cite{ghahramani2015probabilistic,oyedotun2017deep,moore2015talking,sanchez2016comparative,tang2016extreme}.
The construction of intelligent machines in various industries presupposes the existence of symbolic structures, the ability of them to demand and the existence of knowledge (raw material). Once artificial intelligence has intelligence equal to or greater than man's, political and social change will inevitably arise, in which AI has all the advantages of gaining if it realizes that it does not need humans to colonize the universe \cite{swain2014approaching,abadi2016tensorflow,hinton2006fast,szegedy2016rethinking}.
Recent advancement in artificial technology depicts orbiting communications satellites in the space with its 486 processors. In the future, self-replicating artificial intelligence could easily be made with all human colonies outside the earth, and the human race will never be able to fight in the empty space on equal terms \cite{varela2017embodied}.

\section{Knowledge based towards understanding the natural language}

The most significant characteristics of the natural language are that it has the ability to serve as its particular meta-language. It is easily possible to utilize the natural language in order to provide instruction in the use as well as describe the language itself (Weston et al., 2015). It is because, human usually capable to utilize their natural language to describe about natural language itself. The advancement in the natural language understanding significantly provide effective educable cognitive agent role whose task domain contains the understanding of language and whose discourse domain contains the knowledge of their own language.

\section{Spatial frames of reference computation for narrative text understanding}

This AI technique provide the spatial references which explicitly build in order to indicate on how the reference frame should be established and provide the determination of burden of its determination is left to the hearer’s or the reader’s inferences. It’s also intended towards developing heuristics which significantly aid for dynamically computing of reference frames. 
The rule based expert systems; they have important characteristics related to multiple production rules and levels that embody the knowledge of the distinct document image’s characteristics. It’s includes the inference engine which utilizes the knowledge base in order to have identities’ interpretations of several blocks logical within the image \cite{varela2017embodied,chen2016deeplab,chen2016unconstrained}. The control inference engine is applied in several distinct rules levels on the image data for block’s identification. The lowest rule level is the knowledge rule that examines the intrinsic block’s properties. The control rules act and guide the search as the mechanism of focuses on attention. The strategy rules determine whether the constant image interpretation has been achieved or not.


\begin{figure*}
	\centering
	\includegraphics[width=5in,height=3.5in]{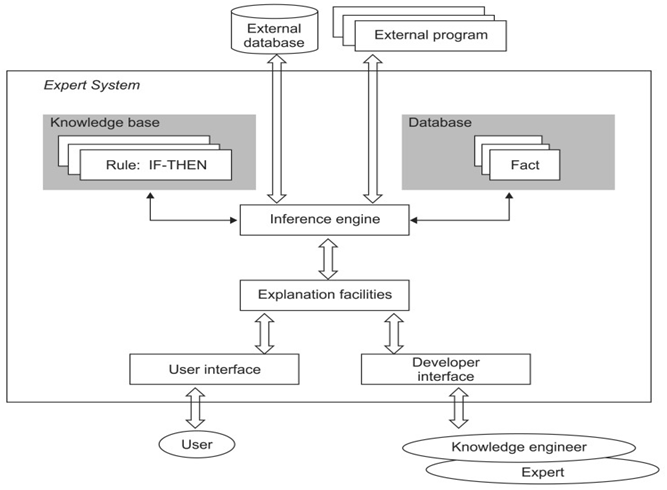}

	\caption{Rule-based expert system structure}
\end{figure*}

The cognitive letter recognition model; it is based on the feature model’s utilization. Within the acquisition time, the letters is subjected towards developing the object oriented quad tree model. In general, all of the acquired letters are integrated into object knowledge, similar quad tree and density values. At the time of recognition, the densities and quad tree analysis of the current letter are related to store quad tree density. 

\section{What are open challenges?}

Challenges of AI technology include the following;
Within near future, the artificial intelligence goals were to affect the society from law and economics in several technical terms including security, verification, validity, and control \cite{varela2017embodied,dong2016image,gongal2015sensors,nguyen2015innovation,du2015hierarchical,tang2013deep}.  The major short-term threats of Artificial Intelligence include the devastating race of arms in fatal autonomous weapons and full dependence of our life on technology will eventually lead to unemployment problem, social discrimination and power inequality in societies. 
long term use of robotics will create a big challenge includes  to human beings as they will take over the world. In addition, with in a given period of time, AI will become better in solving tasks compared to human beings hence lose of jobs. For example drivers will be ruled over by robotic cars, tellers will be out competed by robotic tellers and many others \cite{stoica2017berkeley,sunderhauf2015performance,chen2015deepdriving}. As a result of Artificial Intelligence out competing humans will create a big challenge on human thinking. But in the positive side, the AI technology advancement can also considerably aid for eradicating the disease, war and poverty level.
Rapid research of Artificial intelligence in robotics is reflected as the large existential threats that are faced by the humanity.  The destructive method developed in the artificial intelligent robot can also increase the risk for the super-intelligent system. For example, the ambitious project of geo-engineering can wreak havoc of ecosystem. This considerably increases the concerns about the AI system advancement which is not malevolence but have increased competency. Although, the super-intelligent artificial intelligence tools can considerably aid in fulfilling the goals un-alignment of goals may cause an increasing problem \cite{turan2017non,papernot2016limitations,turan2018deep,turan2018sparse,turan2018unsupervised,turan2017non1,turan2017deep1,turan2017endosensorfusion,huang2016deep,lillicrap2015continuous}. 
Artificial intelligence is simply the next wave of automation, which considerably allows the machines to do tasks that previously required attention and human intelligence. In the short term it can replace people, but above all, it changes the nature of the work that humans do. In the long run, automation creates more and different types of jobs, which is why not everyone has jobs today.  Further, challenges include the large dependence of people towards technology for their work. This considerably increases several psychological, physical and mental issues. In order hand, it considerably reduces the unemployment, economic and power balance and disability. It is believed that until the end of 2021, there will be the beginning of a disruptive tidal wave of artificially intelligent robots in our daily life, such solutions powered by cognitive or artificial intelligence technology will significantly displace the jobs, with the great impact felt in logistics, transportation, consumer services and educational process. Furthermore, it can considerably increase the privacy, security and authenticity issues within the society.

\section{What are solution suggestions?}

To reduce the destructive effects of AI, it is essential to develop the symbolic approach, which should allow us to operate with weakly formalized representations and their meanings. The success and effectiveness of solving new problems depend on the ability to allocate only essential information, which requires flexibility in the methods of abstraction. While a normal program sets one's own way of interpreting the data, which makes its work look prejudiced and purely mechanical. The intellectual task, in this case, solves only the person, the analyst or the programmer, not knowing to entrust this machine. As a result, a single abstraction model is created, a system of constructive essences and algorithms. And flexibility and universality translate into significant resource costs for non-typical tasks, that is, the system from the intellect returns to brute force. Furthermore, the hybrid approach should also be developed in order to provide the synergistic combination of neural and symbolic models achieves a full range of cognitive and computational capabilities. For example, expert rules of inference can be generated by neural networks, and generating rules are obtained through statistical training. Supporters of this approach believe that hybrid information systems will be much stronger than the sum of different concepts separately. Furthermore, to decrease the potential threats of artificial intelligent technology, the rational risk management process that includes the potential principles of adopting the expensive precautions, which have high cost even for the lower probability risks, can considerably provide the benefits (Russell et al., 1995). The global nature of artificial intelligence risks if fails to transfer the ethical goals then it’s absolutely reasonable to estimate the longer-term AI research risks as even larger than those of climate change. It is also essential to adopt effective information system which should be established in order to effectively improve the artificial intelligence safety by initiating the awareness on the expert’s working part on AI, decision-makers, and investors. The risk’s information integrated with AI progress should also have to understandable and accessible for the broad audience. In addition, it is also instigated to adopt the AI safety tools in order to reduce the authentication and security issues. It is also essential to develop an effective global coordination and cooperation system in order to build the competitive environment in which the dangerous race of artificial intelligent arms should be reduced \cite{russell2016artificial,jordan2015machin,lake2015human,marsland2009machine,lenz2015deep,ghahramani2015probabilistic}.
It is believed that although the future of artificial intelligence has a tremendously beneficial impact on the economy and daily lives of Americans still it increases the issues related to privacy, security, unemployment, technological dependency, and authenticity within the society. Therefore, it is essential to develop the preventive control solution for mitigating or reducing such challenges.

\section{What are future predictions for AI?}

There is a great increase in the discussion about the importance of AI in the recent time leading to future discussions about the existence of Artificial Intelligence in the world. The idea of creating AI is aimed at making human life easier.  However, there is still a big debate about advantages and disadvantages of AI in the whole \cite{muller2016future,esteves2003generalized,vedaldi2015matconvnet,eitel2015multimodal,yang2015robot,cirecsan2010deep,lu2015transfer}.
With the introduction and successful implementation of Artificial Intelligence (AI) solutions, many industries in the world are and will benefit from increased profitability and will still have good economic growth rates. In addition, artificial Intelligence opportunities will be aiming at innovative, human centered approaches and measuring the applicability of robotic technology to various industries and companies in the entire world.
Artificial Intelligence will also revolutionize the way different companies in the world grow and compete by representing new production ideas that will derive profitability in businesses \cite{muller2016future,schmidhuber2015deep,brighton2015introducing,bai2015subset,veeriah2015differential}. So as to realize such opportunities, it will require most of the companies in the world to become more active in the development of various Artificial Intelligence strategies such as placing human factors to central nucleus. In addition, they will focus on developing various responsible Artificial Intelligence machines having moral and ethical values which will result into positive results and empowerment of people to do things that they are well versed with.
Construction of various Artificial Intelligence systems will help the entire world to industrial sector to presuppose the available symbolic structures such as, the ability to reason and also knowledge existence. In addition, at the time Artificial Intelligence acquires intelligence greater or equal to that of human beings, there will be a concern about social and political change .In furthermore, AI will have all the advantages of colonize the world without the help of human beings. In the near future, self-replicating AI could be made where human colonies beyond the earth will never have potentials to fight in the free space with critical terms. The future Artificial Intelligence in various regions in the world may be as a result of various investigation technologies such as stellar travel, teleportation and others. 

\section{Which levels will it reach and which issues will be solved in next decades?}

In this case, it is more likely that Artificial Intelligence innovations will strongly emerge in conceivable future. In next decades, the future of AI will be concerned on improving speech, voice, video conferencing and face recognition. Further, Artificial Intelligence will aid for providing the personal assistances and fully automate systems, which will provide assistance in monitoring and surveillance, performing heavy workloads and many others \cite{rooney1983artificial,duan2016benchmarking,levine2016learning,yang2015grasp}. In addition, the future of Artificial Intelligence technology such as robotics will be ensuring self-driven cars, delivery robots and many others. With the great improvement in computer versions and legged locomotion, the robots within environments will become more practical hence helping in agriculture and other service settings. In addition, the robotics will improve on the service delivery hence reducing domestic chores \cite{rooney1983artificial,zhu2017target,cruz2015interactive,vinciarelli2015open,doshi2015deep,wang2015designing,cuayahuitl2015strategic}. Furthermore, as Artificial Intelligence robotics is developing search engines, there will be provision of personal assistance and language gasps by use of mobile devices.  The development of search engines will lead to significant synthesize and improvement of the quality of information. Furthermore, Artificial Intelligence tool will improve the medical and biological system hence reducing the complexity and volume of information challenges concerned human abilities.
Artificial intelligence will be used in the algorithm that materializes various systems and programs. Artificial intelligence will consist of specific hardware and software which intends to   imitate the way of human brain performance.  In general, the areas of Artificial Intelligence application will widely cover both emerging and traditional technologies. Recent research on AI considerably provide the potential effects on organization and industries, for example, the AI technology will be aiming at improving the area of data science to almost 9.6\%, business intelligence to 7.8\%, patient and health care to about 6.3\%, speech recognition 5.3\%, computer vision 5.6\%, improve defense and aerospace system to about 5.3\% and natural language processing to about 5.1\%. The role of Artificial Intelligence tools on product manufacturing operations will lead to toward employment, flexibility and responsive chain of supply. In addition, AI roles will also result to reliable forecasting demands, inventory accuracy and optimization of schedules. The roles of AI will therefore benefit quicker, smarter and environmental efficient process. AI application in security and defense will majorly focus on infrastructure protection. Currently Artificial Intelligence facilitates the power plant, airport  and economic sectors which are quite hard to detect attacks, individual’s anomalous predication of disruption by man-made and natural causes \cite{tirgul2016artificial}.

\begin{figure*}
	\centering
	\includegraphics[width=5.5in,height=3in]{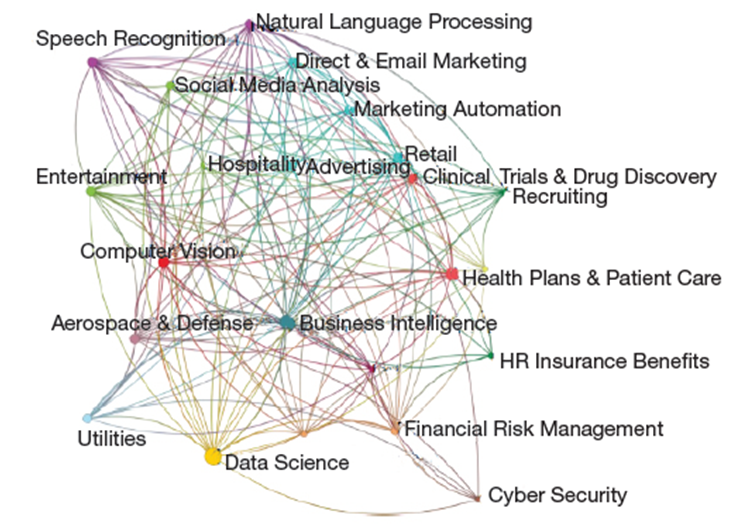}
	\caption{Application areas of AI}
\end{figure*}

Within logistics area, the intervention of Artificial Intelligence will contain efficient vehicles that will be in position to route and make necessary adaptive delivery schedules. In the financial service sector, Artificial Intelligence tools will contain system failure and risk alerts aiming at decreasing malicious attacks various financial systems; such as fraud, market manipulation and reduction in market volatility and trading costs. In the agriculture sector, the intelligent solutions will provide intelligent production mechanisms for processing, consumption, storage and distribution \cite{tirgul2016artificial}. The Artificial solution will also provide given timely data on crops that will involve use of proper materials such as chemicals and fertilizers. Artificial Intelligence will also be used in consumer goods and services so as to  utilize machine learning processes to match consumer demand and enable them get best ;practices at reduced ;prices. In communication sector, there will be improvement in bandwidth and storage and Web translation languages. Within the education sector, Artificial Intelligence solution will intervene basic meaningful adaptive learning basing on adaptive learning complemented by individual learning in the classroom, accurate measurement of student's sensitivity and development of students. Easy manual techniques and judgment will be supplemented by artificial intelligence. The medical \& health-care will provide various health evaluations to patients, decision support for prescribing drugs and indication. Artificial intelligence will be used in large-scale genome researches to determine new drugs, give necessary support for finding new genetic problems, efficiency and safety. The evidence based health and medicine will help various physicians gain confidence which will require supplementary support by patients \cite{gottfredson1998general}.
In customer service sector, AI systems will provide the virtual assistants which will aim at increasing the reproduction and interpretation abilities of human language with greater precision. For example, chat bots redraw the landscape of the IT ecosystem. They will replace themselves and applications, and service personnel in companies, and even entire operating systems. Chat-bot (Chat-bot) - this program will contain an interlocutor, which will be designed to communicate and help people. At the other end, there will be a complex system based on several Artificial Intelligence technologies. Chat bots, oriented to business tasks, will help to can take up best flights, diet, fitness trainings, booking of a hotel, make purchase, that is to say; they will represent a unique sub-sector of assistance and advice. Personal assistants are a kind of incarnation of chat bots, although more common because the technology will be developed by the largest IT companies \cite{turan2017endosensorfusion,mohamed2015variational,zhu2017target,cruz2015interactive,vinciarelli2015open,doshi2015deep,wang2015designing}. Currently, hundreds of millions of people interact with personal digital assistants on platforms such as Google, Apple, Amazon, Facebook and others. This technology with the help of personal assistants and chat-bots will be more effective which will make a great transition from the graphical user interface (GUI) to the Conversational User Interface (CUI) the key trend of the next decades. The predictive algorithms and machine learning based on AI tools will provide sale’s forecasts in specific markets and also effective provision and optimization of inventory as they will help forecast income and determine the necessary quantities of a particular input. In addition, modern Artificial Intelligence systems will control robotics to provide surveillance, security and attacks without threatening the human life in Warfield. AI applications in robotics will have diverse objectives related to automation of military applications, industrial processes and space exploration. The use of AI technology in medicine and surgeries will significantly help in provision of safe work as the machinery occupied will reduce the degree of error that could occur in surgery, avoid a tragic outcome \cite{dilsizian2014artificial}.
The disaster recovery and management application of Artificial Intelligence will considerably remark the provision of remedial and control actions in the aftermath of man-made and environmental disasters. Within disasters, the considerable will optimize the mobile networks and allocate the smart bandwidth \cite{imran2014aidr}. Further, the satellite feed and unmanned drones having image recognition and processing feature will help in assessing damage of infrastructure and provide predictions aimed at avoiding traffic congestion and various structural stability by adopting the adaptive routing system.

\section{Conclusion}

In this way, artificial intelligence can achieve great discoveries and advances for humanity due to its multiple possibilities. Most artificial intelligence systems have the ability to learn, which allows people to improve their performance over time. The adoption of AI outside the technology sector is at an early or experimental stage \cite{lecun2015deep,papernot2016limitation,levine2016end,huang2016deep,turan2017deep,vrigkas2015review, sanchez2016comparative, tang2016extreme, chen2016deeplab, chen2016unconstrained, papernot2016limitations, levine2016end, huang2016deep,turan2018magnetic,yoo2015deep,campos2015diving, russell2015research, das2015performance,salakhutdinov2015learning,schmidhuber2015learning,chen2015road,vrigkas2015review}. The evidence suggests that AI can provide real value to our lives.AI bases its operation on accessing huge amounts of information, processing it, analyzing it and, according to its operation algorithms, executing tasks to solve certain problems. Due to the new computing architectures of the cloud, this technology becomes more affordable for any organization.

\ifCLASSOPTIONcompsoc
  \section*{Acknowledgments}
\else
  \section*{Acknowledgment}
\fi

The authors would like to thank...

\ifCLASSOPTIONcaptionsoff
  \newpage
\fi



%

\bibliographystyle{IEEEtran}
\bibliography{mybibfile}

%





\end{document}